\documentclass[runningheads]{llncs}
\usepackage[T1]{fontenc}
\usepackage{graphicx,verbatim}
\usepackage{booktabs}
\usepackage{multirow}
\usepackage[misc]{ifsym}
\usepackage[final]{microtype}
\usepackage[hyphens]{url}
\begin{document}
\makeatletter
\def\@fnsymbol#1{\ensuremath{\ifcase#1\or \star \or \dagger \or \ddagger \or
   \mathsection \or \mathparagraph \or \| \or \star\star \or \dagger\dagger
   \or \ddagger\ddagger \else\@ctrerr\fi}}
\makeatother
\title{PET-F2I: A Comprehensive Benchmark and Parameter-Efficient Fine-Tuning of LLMs for PET/CT Report Impression Generation}
%
\author{Yuchen Liu\thanks{These authors contributed equally to this work.} \and
Wenbo Zhang\inst{1,4}\protect\footnotemark[1] \and
Yinghao Zhang\inst{1,4} \and 
Liling Peng\inst{2} \and 
Yichi Zhang\inst{1,4} \and 
Yu Fu\inst{3} \and 
Xin Guo\inst{1,4} \and 
Chao Qu\inst{1,4}\protect\footnotemark[2] \and 
Yuan Qi\inst{1,4}\protect\footnotemark[2] \and 
Le Xue\inst{1,4}\thanks{Corresponding authors.}
}
\authorrunning{Yuchen Liu et al.}
%
\institute{Fudan University \and
Shanghai Universal Medical Imaging Diagnostic Center \and 
Lanzhou University \and 
Shanghai Academy of Artificial Intelligence for Science
}


  
\maketitle              
\begin{abstract}
PET/CT imaging is pivotal in oncology and nuclear medicine, yet summarizing complex findings into precise diagnostic impressions is labor-intensive. While LLMs have shown promise in medical text generation, their capability in the highly specialized domain of PET/CT remains underexplored. We introduce \textbf{PET-F2I-41K} (PET Findings-to-Impression Benchmark), a large-scale benchmark for PET/CT impression generation using LLMs, constructed from over 41k real-world reports. Using PET-F2I-41K, we conduct a comprehensive evaluation of 27 models across proprietary frontier LLMs, open-source generalist models, and medical-domain LLMs, and we develop a domain-adapted 7B model (\textbf{PET-F2I-7B}) fine-tuned from Qwen2.5-7B-Instruct via LoRA. Beyond standard NLG metrics (e.g., BLEU-4, ROUGE-L, BERTScore), we propose three clinically grounded metrics---\emph{Entity Coverage Rate} (ECR), \emph{Uncovered Entity Rate} (UER), and \emph{Factual Consistency Rate} (FCR)---to assess diagnostic completeness and factual reliability. Experiments reveal that neither frontier nor medical-domain LLMs perform adequately in zero-shot settings. In contrast, PET-F2I-7B achieves substantial gains (e.g., 0.708 BLEU-4) and a 3.0$\times$ improvement in entity coverage over the strongest baseline, while offering advantages in cost, latency, and privacy. Beyond this modeling contribution, PET-F2I-41K establishes a standardized evaluation framework to accelerate the development of reliable and clinically deployable reporting systems for PET/CT.


\end{abstract}

\section{Introduction}
\begin{figure}[t]
\centering
\includegraphics[width=\columnwidth]{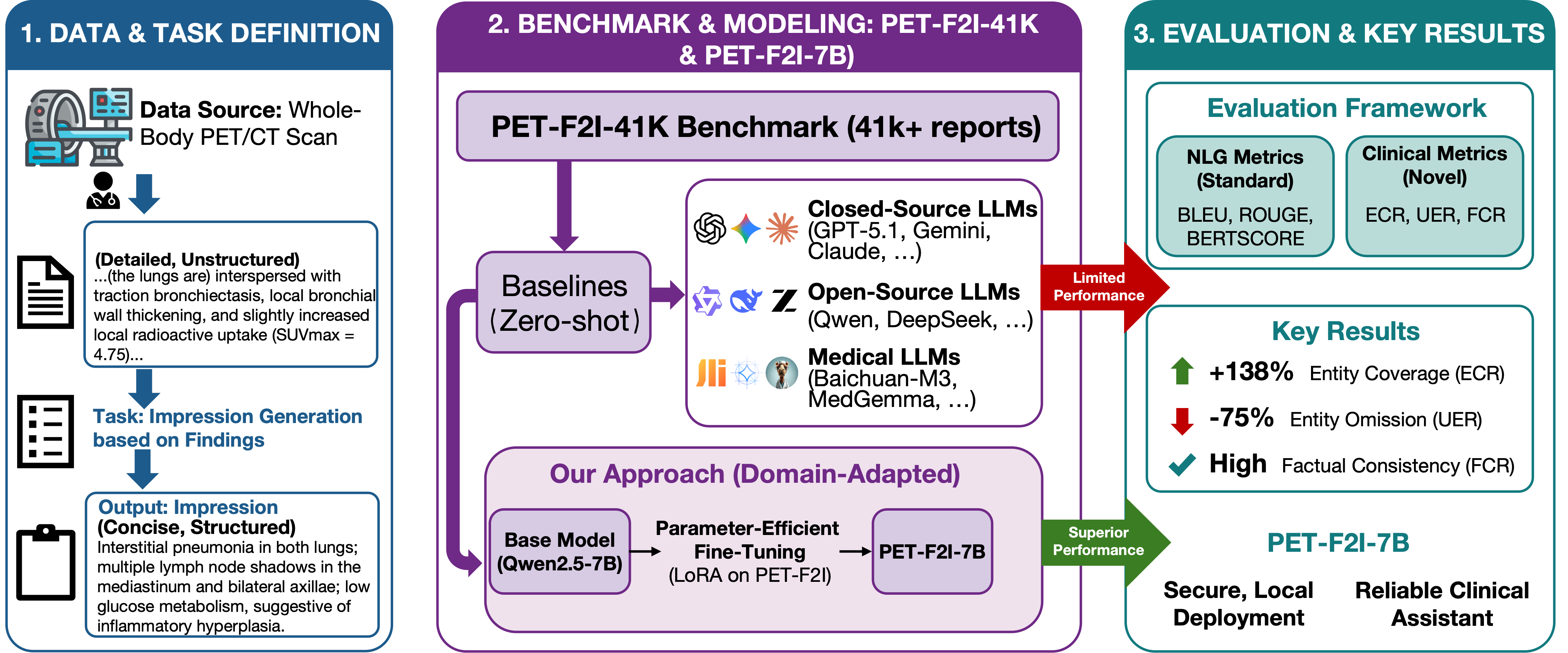} 
\caption{\textbf{Overview of the PET-F2I-41K framework.} The pipeline encompasses clinical task formalization, comprehensive benchmarking of 27 LLMs, and multi-dimensional evaluation, demonstrating the absolute superiority of our domain-adapted PET-F2I-7B (+138\% ECR, -75\% UER).}
\label{fig:framework}
\end{figure}

Positron emission tomography/computed tomography (PET/CT) is an indispensable imaging modality in modern oncology. However, synthesizing intricate radiological findings into concise diagnostic impressions remains a severe clinical bottleneck. This cognitive process is profoundly time-intensive, demands subspecialty expertise, and is highly susceptible to inter-reader variability under escalating clinical workloads.
While Large Language Models (LLMs) show promise in medical text generation, PET/CT reporting presents unique subspecialty challenges \cite{zhang2026pet2rep}. 
It demands strict formatting and specialized terminology, such as standardized uptake value (SUV) and tumor-node-metastasis (TNM) staging. Crucially, generative errors like hallucinated or omitted malignancies directly threaten patient safety. Furthermore, deploying proprietary frontier LLMs faces critical barriers: high latency, operational costs, and strict patient health information (PHI) privacy regulations render cloud-based solutions unviable for clinical workflows. A critical impediment in this domain is the absence of clinically rigorous evaluation frameworks \cite{xue2025petwb}. To bridge this gap, we introduce PET-F2I-41K, the first large-scale benchmark engineered for PET/CT impression generation. Because conventional metrics (e.g., BLEU, ROUGE) fail to capture fatal clinical inaccuracies, we propose three novel, clinically grounded metrics: Entity Coverage Rate (ECR), Uncovered Entity Rate (UER), and Factual Consistency Rate (FCR). 
As outlined in Fig.~\ref{fig:framework}, evaluating 27 models on PET-F2I-41K reveals a critical insight: neither massive frontier LLMs nor specialized medical systems (whether broadly pre-trained or QA-tuned) can meet the rigorous structural and cognitive demands of PET/CT reporting. Instead, precise domain adaptation guarantees robust clinical efficacy.
\textbf{Our core contributions are:} \textbf{(1)} We establish \textbf{PET-F2I-41K}, the first large-scale benchmark (41,191 reports) for this task, alongside three novel clinical metrics (ECR, UER, FCR) to rigorously evaluate diagnostic completeness and factual consistency. \textbf{(2)} We comprehensively benchmark 27 LLMs, exposing the clinically unacceptable entity omission rates of massive frontier and generalized medical models in zero-shot settings. \textbf{(3)} We propose \textbf{PET-F2I-7B}, a parameter-efficient, locally deployable architecture that sets a new state-of-the-art, delivering a 3.0$\times$ improvement in exact entity coverage over the strongest baseline while ensuring strict PHI privacy.


\section{Related Work}

While LLMs excel in general clinical summarization \cite{singhal2023medpalm,singhal2023medpalm2}, radiology research predominantly targets chest radiograph report generation using retrieval-augmented or vision-language models \cite{yu2023radiology,chen2020r2gen,chen2022r2gencmn,liu2023radiology,li2024llava,zhang2023biomedclip,moor2023medflamingo}. However, PET/CT impression generation diverges fundamentally, demanding complex multi-region integration and highly specific tracer terminology. Consequently, whether adapted via broad biomedical pretraining or medical instruction tuning, specialized medical LLMs (e.g., BioGPT \cite{luo2022biogpt}, Meditron \cite{chen2023meditron}, Med-PaLM 2 \cite{singhal2023medpalm2}) exhibit limited transferability and remain vulnerable to clinical hallucinations \cite{ji2023hallucination,umapathi2023med}.
Furthermore, conventional lexical metrics (e.g., BLEU, ROUGE) inadequately detect clinically severe errors like critical omissions or fabricated diagnoses. Drawing on clinical entity extraction principles \cite{jain2021radgraph,smit2020chexbert}, we introduce targeted metrics to explicitly quantify diagnostic completeness and factual fidelity. To address the deployment barriers of massive models, we utilize parameter-efficient fine-tuning \cite{hu2022lora,dettmers2023qlora}. By selectively adapting a compact 7B-parameter architecture, we demonstrate that targeted domain adaptation successfully overcomes the limitations of generalized and medical-domain LLMs, establishing a highly accurate, locally deployable paradigm.

\section{Methods}

\begin{figure}[t]
  \centering
  \includegraphics[width=\columnwidth]{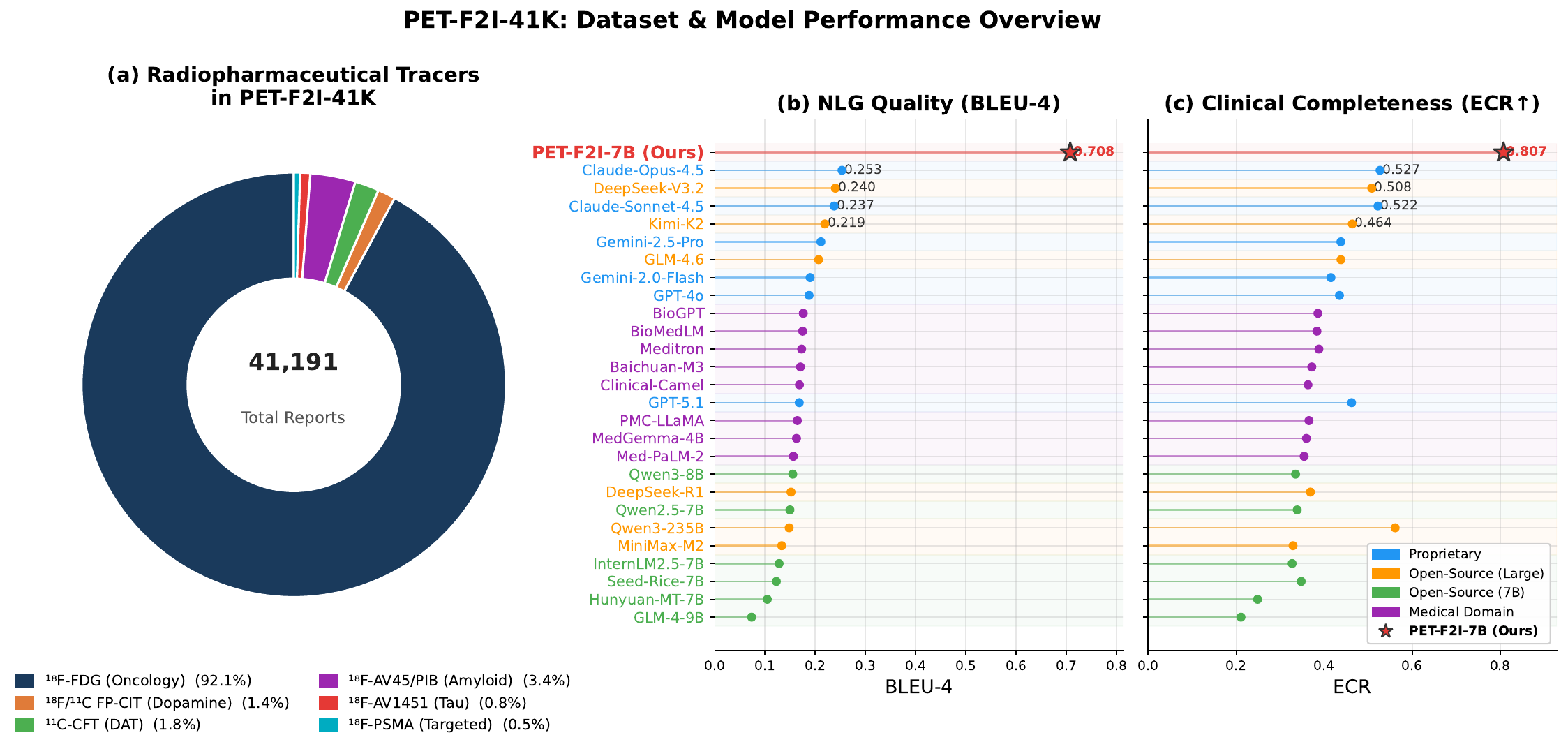}
  \caption{(a)~Tracer distribution in PET-F2I-41K (41{,}191 reports). (b,\,c)~27 models ranked by BLEU-4 and ECR; PET-F2I-7B ($\star$) consistently outperforms the performance of frontier proprietary, large-scale open-source, and specialized medical LLMs.}
  \label{fig:composite}
\end{figure}
\subsection{Dataset Curation and Benchmark Construction}
We introduce PET-F2I-41K, the first large-scale corpus for PET/CT impression generation, comprising 41,191 real-world reports (2013--2023). To rigorously prevent data leakage, we applied a strict patient-level split: 40,691 training, 500 validation, and 500 test samples. The cohort (mean age 58.3; 56.7\% male) spans a diverse clinical spectrum. As illustrated in Fig.~\ref{fig:composite}(a), while predominantly $^{18}$F-FDG, the dataset incorporates critical minority tracers for neurological and targeted molecular imaging. The extreme linguistic density (\emph{Findings} avg.\ 870 characters; \emph{Impressions} avg.\ 240 characters) reflects the rigorous cognitive demands of clinical practice. To systematically evaluate this complexity, we constructed a comprehensive benchmark assessing 27 LLMs. As demonstrated in the resulting leaderboards (Fig.~\ref{fig:composite}(b), (c)), proprietary frontier, large-scale open-source, and even specialized medical baselines struggle with clinical completeness, whereas our domain-adapted PET-F2I-7B establishes absolute superiority, substantiating the critical need for task-specific adaptation

\subsection{Models and Optimization Strategy}
We benchmark 27 LLMs across frontier proprietary (e.g., GPT-5.1~\cite{openai2023gpt4}, Claude Opus~\cite{anthropic2024claude}, Gemini-2.5~\cite{comanici2025gemini}), open-source (e.g., Qwen~\cite{qwen2024qwen2}, DeepSeek~\cite{deepseek2024v2}, GLM~\cite{glm2024chatglm}), and specialized medical domains. Addressing clinical privacy and efficiency imperatives, we propose \textbf{PET-F2I-7B}, a domain-adapted model fine-tuned from Qwen2.5-7B-Instruct via LoRA~\cite{hu2022lora}. By applying LoRA modules ($r=64, \alpha=128$) to all linear projections, we optimize via AdamW (peak LR $1.0 \times 10^{-4}$, 3 epochs, 2048-token context) across 2 RTX 4090 GPUs. This parameter-efficient approach enables seamless local inference on one RTX 4090, intrinsically securing patient health information by preventing external server transmissions. 

\subsection{Evaluation Framework and Clinical Metrics}
For automatic text evaluation, we compute standard metrics including BLEU~\cite{papineni2002bleu}, ROUGE~\cite{lin2004rouge}, METEOR~\cite{banerjee2005meteor}, BERTScore~\cite{zhang2019bertscore}, and SBERT~\cite{reimers2019sentence}. However, standard lexical metrics inherently fail to capture clinically catastrophic errors, such as the critical omission of malignant lesions or the hallucination of non-existent diagnoses. To rigorously assess clinical viability, we constructed a comprehensive medical lexicon aggregating the THUOCL dictionary, extensive anatomical databases, and a manually curated repository of specialized PET/CT vernacular. Utilizing a greedy longest-match named entity recognition (NER) framework over this dictionary, we propose three novel, clinically grounded metrics.
Let $E_{ref}$ and $E_{gen}$ denote the sets of extracted clinical entities in the reference and generated impressions, respectively. We define the \textbf{Entity Coverage Rate (ECR)} to quantify diagnostic completeness as the proportion of reference entities successfully reproduced:
\begin{equation}
ECR = \frac{|E_{ref} \cap E_{gen}|}{|E_{ref}|}
\end{equation}
Conversely, to explicitly quantify the incidence of clinical hallucinations, we introduce the \textbf{Unsupported Entity Rate (UER)}. This metric calculates the proportion of generated entities absent from the source findings, serving as a strict proxy for diagnostic fabrication:
\begin{equation}
UER = \frac{|E_{gen} \setminus E_{ref}|}{|E_{gen}|}
\end{equation}
Finally, practical clinical utility demands strict adherence to institutional reporting structures. We define the \textbf{Format Compliance Rate (FCR)} to evaluate the structural integrity of the generated output. Let $C = \{c_1, \dots, c_N\}$ represent $N=5$ predefined formatting criteria (e.g., numerical sectioning, anatomical markers, terminology density). For each criterion $c_i$, we compute a rule-based compliance score $s_i \in \{0, 0.5, 1\}$. The FCR is formally defined as the average compliance score:
\begin{equation}
FCR = \frac{1}{N} \sum_{i=1}^{N} s_i
\end{equation}
Through the integration of ECR, UER, and FCR, alongside tracer-stratified analyses, our benchmark establishes a multi-dimensional paradigm that strictly measures genuine diagnostic utility over superficial textual overlap.

\begin{table*}[t]
\centering
\renewcommand{\arraystretch}{1.1}
\caption{\textbf{Quantitative evaluation of 27 LLMs on the PET-F2I-41K benchmark.} \textbf{Bold} and \underline{underline} denote the best overall and best-in-category scores, respectively. The domain-adapted PET-F2I-7B demonstrates absolute superiority across all standard NLG and proposed clinical metrics, particularly maximizing exact entity coverage (ECR).}
\label{tab:main_results}
\resizebox{\textwidth}{!}{%
\begin{tabular}{l|l|ccccc|ccc}
\toprule
\textbf{Category} & \textbf{Model} & \textbf{BLEU-4}$\uparrow$ & \textbf{ROUGE-L}$\uparrow$ & \textbf{METEOR}$\uparrow$ & \textbf{BERTScore}$\uparrow$ & \textbf{SBERT}$\uparrow$ & \textbf{ECR}$\uparrow$ & \textbf{UER}$\downarrow$ & \textbf{FCR}$\uparrow$ \\
\midrule
\multirow{6}{*}{Proprietary} 
 & Claude Opus 4.5 \cite{anthropic2024claude}& \underline{0.2532} & 0.3520 & \underline{0.5692} & 0.7590 & 0.6433 & \underline{0.5268} & 0.5417 & 0.9814 \\
 & Claude Sonnet 4.5 \cite{anthropic2024claude}& 0.2374 & 0.3587 & 0.5369 & 0.7562 & 0.6286 & 0.5222 & 0.5434 & 0.9734 \\
 & Gemini 2.5 Pro \cite{comanici2025gemini}& 0.2112 & 0.4193 & 0.5127 & \underline{0.7643} & \underline{0.6597} & 0.4379 & 0.5152 & 0.9618 \\
 & Gemini 2.0 Flash \cite{team2023gemini}& 0.1897 & \underline{0.4792} & 0.4653 & 0.7516 & 0.6401 & 0.4153 & \underline{0.4947} & 0.9628 \\
 & GPT-4o \cite{openai2023gpt4}& 0.1876 & 0.3248 & 0.4577 & 0.7344 & 0.5990 & 0.4346 & 0.5883 & 0.9756 \\
 & GPT-5.1 \cite{openai2023gpt4}& 0.1681 & 0.2963 & 0.4768 & 0.7327 & 0.5888 & 0.4622 & 0.6250 & \textbf{0.9898} \\
\midrule
\multirow{6}{*}{Open-Source (Large)} 
 & DeepSeek V3.2 \cite{deepseek2024v2}& \underline{0.2401} & \underline{0.3822} & \underline{0.5296} & \underline{0.7566} & \underline{0.6339} & 0.5080 & 0.5704 & \underline{0.9778} \\
 & Kimi-K2 \cite{team2025kimi}& 0.2190 & 0.3398 & 0.5052 & 0.7557 & 0.6152 & 0.4638 & \underline{0.5062} & 0.9548 \\
 & GLM-4.6 \cite{glm2024chatglm}& 0.2067 & 0.3639 & 0.4872 & 0.7529 & 0.6139 & 0.4381 & 0.5329 & 0.9714 \\
 & DeepSeek R1 \cite{deepseek2024v2}& 0.1517 & 0.2334 & 0.3805 & 0.7003 & 0.5489 & 0.3686 & 0.6120 & 0.9024 \\
 & Qwen3-235B \cite{yang2025qwen3}& 0.1480 & 0.2827 & 0.4201 & 0.7322 & 0.6194 & \underline{0.5612} & 0.6709 & 0.9702 \\
 & MiniMax-M2 & 0.1332 & 0.2323 & 0.3588 & 0.6875 & 0.5697 & 0.3292 & 0.6681 & 0.8590 \\
\midrule
\multirow{6}{*}{Open-Source (7B)} 
 & Qwen3-8B \cite{yang2025qwen3}& \underline{0.1552} & \underline{0.3365} & \underline{0.3840} & 0.7076 & \underline{0.7457} & 0.3349 & 0.6464 & 0.9368 \\
 & Qwen2.5-7B \cite{qwen2024qwen2}& 0.1495 & 0.3152 & 0.3739 & \underline{0.7234} & 0.5854 & 0.3387 & 0.6567 & 0.9522 \\
 & InternLM2.5-7B & 0.1280 & 0.2887 & 0.2618 & 0.7029 & 0.5854 & 0.3271 & 0.6833 & 0.9452 \\
 & Seed-Rice-7B \cite{yang2025seedllm}& 0.1225 & 0.2987 & 0.2371 & 0.7010 & 0.6076 & \underline{0.3477} & 0.7207 & \underline{0.9727} \\
 & Hunyuan-MT-7B \cite{zheng2025hunyuan}& 0.1045 & 0.2603 & 0.2688 & 0.6889 & 0.5860 & 0.2487 & 0.6836 & 0.8214 \\
 & GLM-4-9B \cite{glm2024chatglm}& 0.0733 & 0.2189 & 0.2865 & 0.6929 & 0.6286 & 0.2110 & \underline{0.5898} & 0.7178 \\
\midrule
\multirow{8}{*}{Medical Domain} 
 & BioGPT \cite{luo2022biogpt}& \underline{0.1761} & 0.3294 & \underline{0.4323} & 0.7348 & 0.6061 & 0.3857 & 0.5609 & 0.9470 \\
 & BioMedLM \cite{bolton2024biomedlm}& 0.1750 & 0.3305 & 0.4295 & \underline{0.7365} & 0.6027 & 0.3834 & 0.5578 & \underline{0.9530} \\
 & Meditron \cite{chen2023meditron}& 0.1728 & 0.3299 & 0.4253 & 0.7364 & 0.6039 & \underline{0.3879} & 0.5504 & 0.9482 \\
 & Baichuan-M3 \cite{wang2024baichuanm1}& 0.1705 & \underline{0.3359} & 0.4248 & 0.7359 & \underline{0.6080} & 0.3719 & 0.5491 & 0.9400 \\
 & Clinical Camel \cite{toma2023clinical}& 0.1685 & 0.3227 & 0.4201 & 0.7346 & 0.6006 & 0.3631 & 0.5583 & 0.9404 \\
 & PMC-LLaMA \cite{wu2024pmcllama}& 0.1643 & 0.3310 & 0.4187 & 0.7332 & 0.5964 & 0.3653 & 0.5542 & 0.9360 \\
 & MedGemma-4B \cite{yang2024medgemma}& 0.1626 & 0.3345 & 0.4133 & 0.7348 & 0.5967 & 0.3597 & \underline{0.5481} & 0.9328 \\
 & Med-PaLM 2 \cite{singhal2023medpalm}& 0.1563 & 0.3222 & 0.4032 & 0.7298 & 0.5854 & 0.3544 & 0.5484 & 0.9268 \\
\midrule
\textbf{Ours} & PET-F2I-7B & \textbf{0.7075} & \textbf{0.8673} & \textbf{0.8942} & \textbf{0.9103} & \textbf{0.9649} & \textbf{0.8074} & \textbf{0.1649} & \underline{0.9420} \\
\bottomrule
\end{tabular}
}
\end{table*}
\section{Results}

\begin{figure}[t]
  \centering
  \includegraphics[width=\columnwidth]{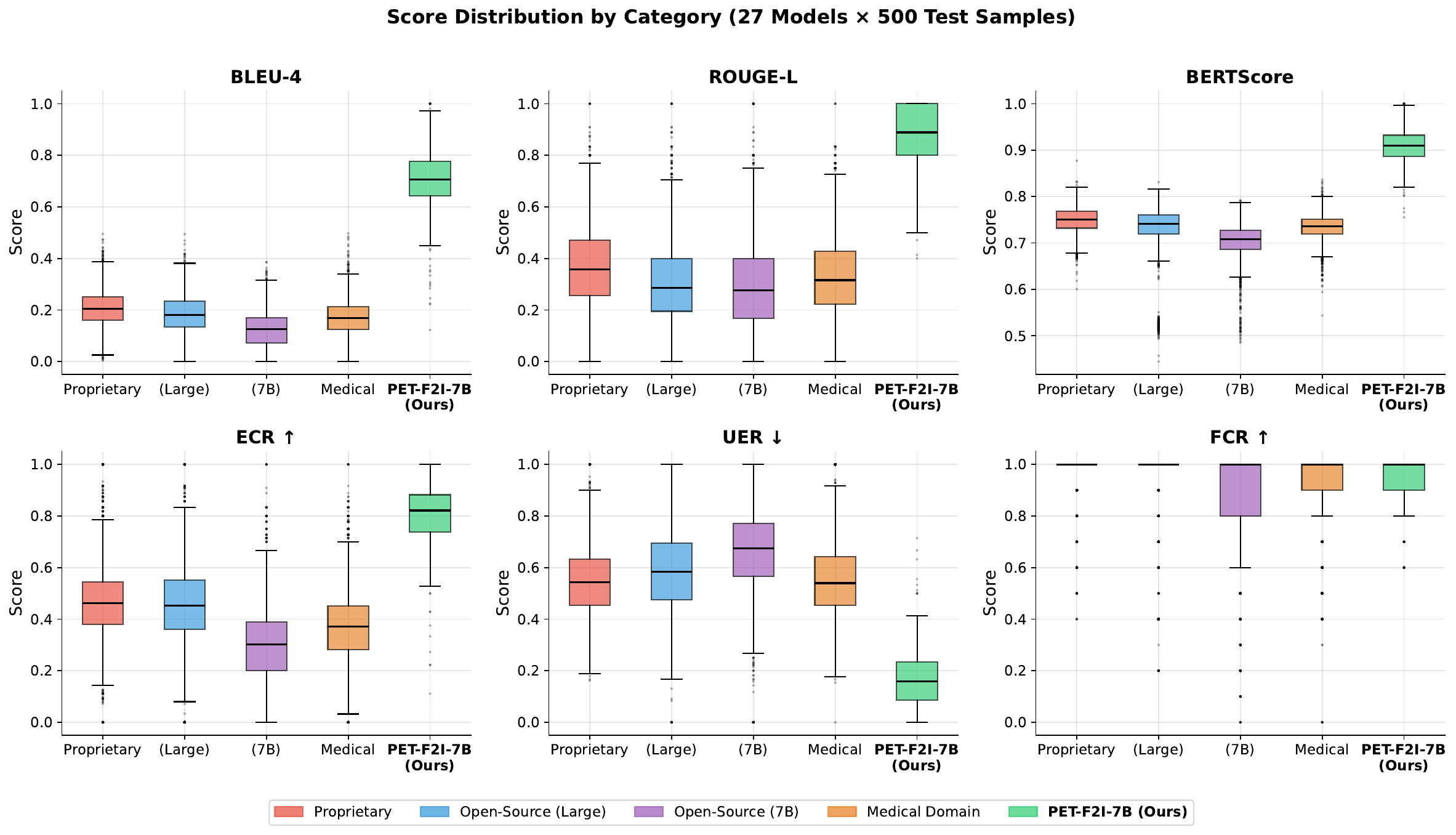}
  \caption{\textbf{Score distributions across model categories ($N=500$).} Beyond superior averages, PET-F2I-7B demonstrates exceptional clinical stability. Crucially, PET-F2I-7B's lowest quartile for exact entity coverage (ECR) strictly dominates the highest quartiles of all zero-shot baselines, underscoring its superior reliability for real-world deployment.}
  \label{fig:boxplots}
\end{figure}

\begin{figure}[t]
  \centering
  \includegraphics[width=\columnwidth]{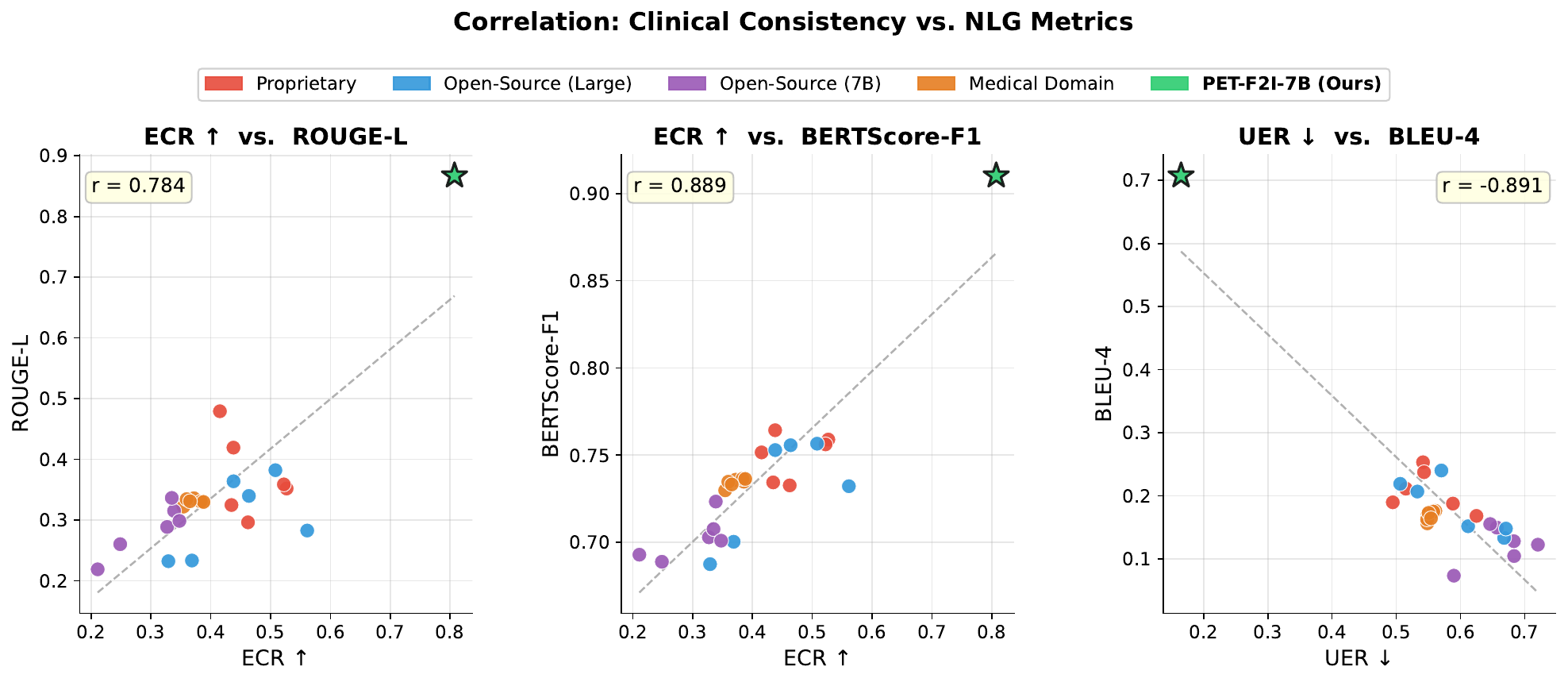}
\caption{\textbf{Correlation between NLG and clinical metrics.} Despite strong macro-level correlations (e.g., ECR vs. BERTScore-F1: $r=0.889$), substantial intra-distribution variance reveals that high lexical overlap does not guarantee clinical accuracy. Models with identical NLG scores frequently exhibit drastically different entity omission rates, proving traditional metrics are insufficient proxies for diagnostic safety and highlighting the necessity of the PET-F2I framework.}

  \label{fig:clinical_nlg_scatter}
\end{figure}
\subsection{Comprehensive Benchmark Analysis}
Table~\ref{tab:main_results} details the evaluation of 27 models on the PET-F2I-41K benchmark. These zero-shot evaluations reveal three critical insights:
First, model scale does not guarantee clinical efficacy. Frontier systems like GPT-5.1 and Claude Opus 4.5 yield suboptimal Entity Coverage Rates (ECR of 0.46 and 0.53, respectively), demonstrating that massive parameter counts cannot inherently resolve PET/CT's terminological complexities. 
Second, generalized medical knowledge offers negligible advantages. Specialized medical LLMs (whether broadly pre-trained or instruction-tuned) perform comparably to generic baselines, proving that broad biomedical exposure fails to transfer to nuanced subspecialty reporting. 
Third, the structural proficiency of baseline models is highly deceptive. Despite maintaining FCRs $>0.90$ by mimicking standard report formats, this superficial accuracy masks profound diagnostic errors. Specifically, these models systematically omit critical findings (low ECR) and fabricate unsupported diagnoses (high UER), rendering them clinically unsafe without targeted adaptation.
Ultimately, our domain-adapted PET-F2I-7B fundamentally significantly outperforms these baselines (Fig.~\ref{fig:boxplots}), achieving absolute superiority and remarkable clinical stability across diverse patient samples.


\begin{figure}[t]
  \centering
  \includegraphics[width=\columnwidth]{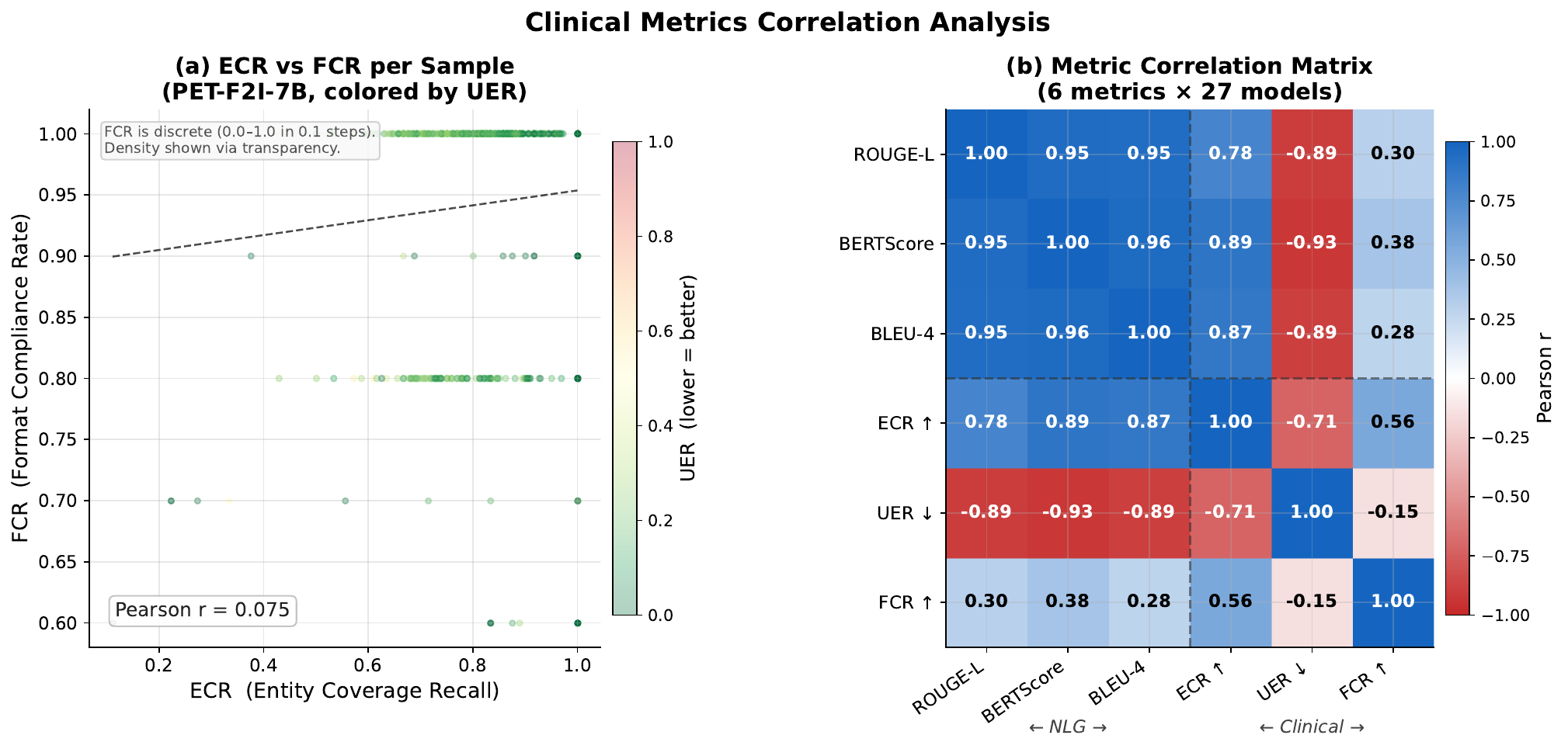}
\caption{\textbf{Independence of clinical metrics and divergence from NLG scores.} \textbf{(a)} Sample-level analysis demonstrates that entity coverage (ECR) and formatting compliance (FCR) are fundamentally orthogonal ($r=0.075$). \textbf{(b)} The correlation matrix reveals that standard lexical metrics exhibit near-zero correlation with clinical factuality (e.g., BLEU-4 vs.\ FCR: $r=0.28$), highlighting their inadequacy for safety evaluation.}
  \label{fig:correlation}
\end{figure}

\begin{figure}[ht!]
  \centering
  \includegraphics[width=\columnwidth]{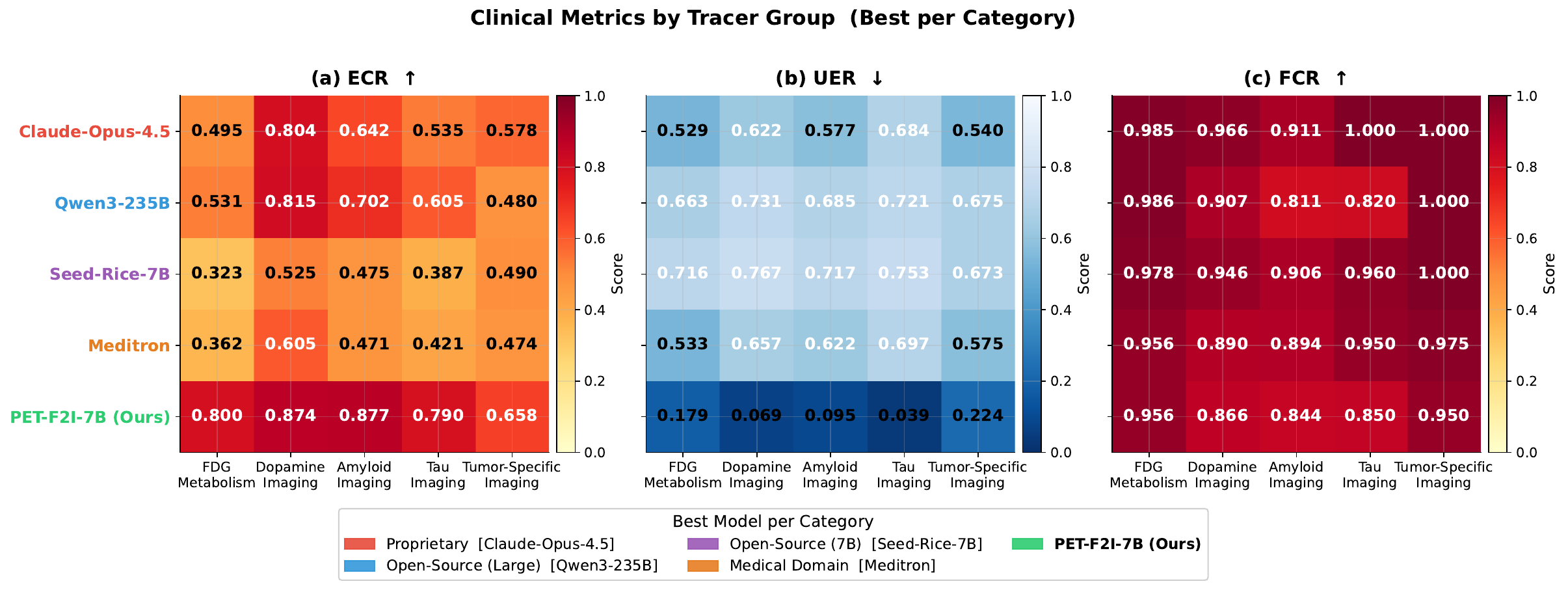}
  \caption{\textbf{Robustness across radiopharmaceutical tracers.} Heatmaps of clinical metrics (rows: models, columns: tracers) reveal that PET-F2I-7B generalizes exceptionally well, maintaining high diagnostic accuracy even on minority tracer profiles.}
  \label{fig:tracer_heatmap}
\end{figure}

\subsection{Clinical Alignment and Generalization Capability}
PET-F2I addresses the fundamental inadequacy of conventional NLG metrics as proxies for clinical safety. While macro-level correlations exist (e.g., ECR vs.\ BERTScore-F1: $r=0.889$), substantial intra-distribution variance reveals that models with high lexical overlap still critically omit diagnostic entities (Fig.~\ref{fig:clinical_nlg_scatter}). More alarmingly, NLG metrics are entirely blind to structural formatting (e.g., BLEU-4 vs.\ FCR: $r=0.28$; Fig.~\ref{fig:correlation}(b)). Sample-level analysis (Fig.~\ref{fig:correlation}(a)) further confirms that entity coverage (ECR) and factual compliance (FCR) are mathematically orthogonal ($r=0.075$). This independence rigorously validates our tripartite design: capturing all findings (ECR), preventing fabrications (UER), and enforcing structural reporting formats (FCR).
Beyond metric alignment, we evaluated model robustness across radiopharmaceutical tracer groups (Fig.~\ref{fig:tracer_heatmap}). Notably, despite an FDG-dominated training corpus, PET-F2I-7B maintains high diagnostic accuracy on out-of-distribution minority scans (e.g., dopamine, amyloid, and tau imaging). This zero-shot generalization demonstrates that PET-F2I-7B internalizes fundamental radiological reasoning rather than merely memorizing common terminological distributions.

\section{Conclusion}
We introduce PET-F2I-41K, the first comprehensive benchmark and clinical evaluation framework for PET/CT impression generation. Evaluating 27 LLMs reveals that zero-shot generalization frequently fails in specialized radiology, suffering from severe entity omissions and hallucinations. To resolve this, our domain-adapted PET-F2I-7B achieves a 3.0$\times$ improvement in exact entity coverage over the strongest zero-shot baseline while enabling secure, privacy-preserving local deployment. Ultimately, precise domain adaptation delivers significantly higher diagnostic safety than arbitrary model scaling or generalized medical pretraining. Future work will address current limitations by expanding to multi-centric, multilingual cohorts, developing multimodal architectures that fuse raw PET/CT images with text, and integrating radiologist preference alignment (e.g., RLHF) to further fortify real-world clinical safety.
    


%
%
%
%
\clearpage
\bibliographystyle{splncs04}
\bibliography{references_corrected}
%
%
%
%
%
\end{document}